\pdfoutput=1

\documentclass[11pt]{article}

\usepackage[]{ACL2023}

\usepackage{times}
\usepackage{latexsym}

\usepackage[T1]{fontenc}

\usepackage[utf8]{inputenc}

\usepackage{microtype}

\usepackage{inconsolata}
\usepackage{hyperref}
\usepackage{enumitem}
\usepackage{tikz}
\newcommand*\circled[1]{\tikz[baseline=(char.base)]{
            \node[shape=circle,draw,inner sep=2pt] (char) {#1};}}

\usepackage{newunicodechar}
\newunicodechar{ֿ}{}

\usepackage{amsmath}
\usepackage{booktabs}

\newcommand{\llamaoneb}{\texttt{Llama-3.2-1B}}
\newcommand{\llamathreeb}{\texttt{Llama-3.2-3B}}
\newcommand{\llamaeightb}{\texttt{Llama-3.1-8B}}
\newcommand{\llamaeightbinstruct}{\texttt{Llama-3.1-8B-Instruct}}
\newcommand{\mistralsevenb}{\texttt{Mistral-7B-v0.1}}
\newcommand{\qwensevenb}{\texttt{Qwen2.5-7B}}

\newcommand{\tuned}{\texttt{informativeness}-tuned}

\title{InFact: Informativeness Alignment for Improved LLM Factuality}

\author{Roi Cohen \\
         Hasso Plattner Institute \\
         University of Potsdam\\
         Germany \\
        \texttt{roi.cohen@hpi.de} \\\And
  Russa Biswas \\
  Dept. of Computer Science \\
Aalborg University \\ Copenhagen, Denmark\\
  \texttt{rubi@cs.aau.dk} \\\And
  Gerard de Melo \\
   Hasso Plattner Institute
  \\ University of Potsdam\\
   Germany \\
  \texttt{gerard.demelo@hpi.de}}

\begin{document}
\maketitle

\begin{abstract}
Factual completeness is a general term that captures how detailed and informative a factually correct text is. For instance, the factual sentence \emph{``Barack Obama was born in the United States''} is factually correct, though less informative than the factual sentence \emph{``Barack Obama was born in Honolulu, Hawaii, United States''}. 
Despite the known fact that LLMs tend to hallucinate and generate factually incorrect text, they might also tend to choose to generate factual text that is indeed factually correct and yet less informative than other, more informative choices. In this work, we tackle this problem by proposing an informativeness alignment mechanism. This mechanism takes advantage of recent factual benchmarks to propose an informativeness alignment objective. This objective prioritizes answers that are both correct and informative. A key finding of our work is that when training a model to maximize this objective or optimize its preference, we can improve not just informativeness but also factuality.
\footnote{We release our code and data at 
\url{https://github.com/roi-hpi/informativeness-alignment}.}

\end{abstract}

\begin{figure}
\setlength{\belowcaptionskip}{-10pt}
    \centering
    \includegraphics[width=0.5\textwidth]{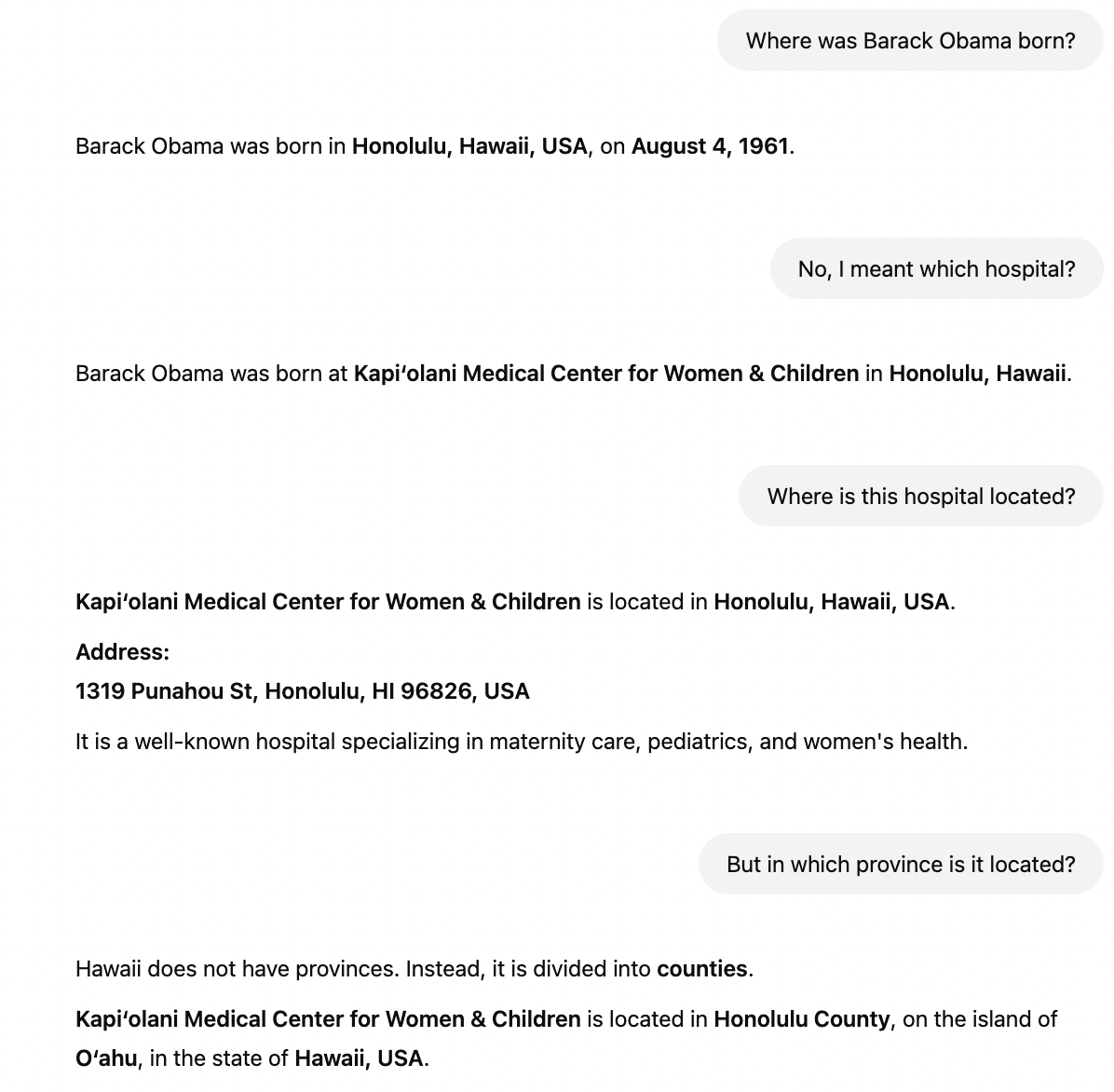}
    \caption{Illustration that an LLM  
    knows a more informative answer than its initial answer.}
    \label{figure:intro}
\end{figure}

\section{Introduction}

Large language models (LLMs) are known to capture and store extensive amounts of factual knowledge \cite{petroni-etal-2019-language, brown2020language, roberts2020much, cohen-etal-2023-crawling, LLMsKGs2023}, as they are trained on vast quantities of text, which includes a significant body of factual knowledge. However, they often hallucinate or generate factually incorrect text \cite{maynez-etal-2020-faithfulness, devaraj-etal-2022-evaluating, tam-etal-2023-evaluating, kaddour2023challenges, huang2024survey}.

Although the way LLMs represent their knowledge remains unclear \cite{Rai2024APR}, it can be effectively accessed via prompting \cite{Veseli2023EvaluatingLM}. For example, modern LLMs are likely to correctly complete the input prompt: \emph{``Barack Obama was born in''}, with multiple different answers -- \emph{United States}, \emph{Hawaii} and \emph{Kapiolani Medical Center for Women \& Children}. While these answers are all factually correct, they differ greatly in specificity and significantly in the level of their informativeness. This highlights an important gap: even when LLMs do not hallucinate, they often fail to provide the most informative answer. Therefore, in this work, we focus beyond the factual correctness of the LLMs to examine the informativeness by evaluating how well the correct entities are represented in the LLM outputs. An answer is considered highly informative if it includes either the most specific entity or all the correct entities.\footnote{In this study, we assume that, unless stated otherwise, the most informative answer that is also fully correct is preferred. While other preferences may be chosen for different users or applications, we argue that to produce increasingly reliable, trustworthy, and knowledgeable models it is advantageous to possess both maximal informativeness and factuality.} 

Our main assumption and motivation for this work is that an LLM might indeed have correct and informative parametric knowledge about a given query, yet generate a less informative answer -- as illustrated in Figure~\ref{figure:intro}. This might occur due to statistical or even spurious correlations between the input text (prompts or questions) and the correct yet less informative answer. 
For example, \emph{``Barack Obama''} and \emph{``The United States''} is more likely substantially co-occur higher than \emph{``Barack Obama''} and \emph{``Kapiolani Medical Center for Women \& Children''} in the training data, leading to the LLM to prefer less specific but more frequent answer.

Therefore, we aim to align the model to prefer the most \emph{informative} or specific answer the LLM knows. Additionally, we also address the well-known existing factual precision problem of the LLMs \cite{augenstein2023factuality} by incorporating this into our alignment mechanism.  

To this end, we begin by formulating the \textit{informativeness-evaluation task} 
and propose a novel framework to create a general informativeness dataset. 
In this dataset, each question or input text is paired with a set of answers associated with a hierarchical set of labels, representing different levels of informativeness. This dataset structure, which includes the informativeness metadata for each label is incorporated directly into the training. 

Building on this dataset, we propose the novel training framework \emph{InFACT}, to align a pre-trained or instruction-tuned LLM model to generate more informative and complete facts. This procedure consists of two stages -- \emph{Structure Tuning} and \emph{Informativeness-Alignment}. \emph{Structure Tuning} phase aims to teach the model to consider the informativeness problem setup as well as to abstain from generating misinformation, whereas the \emph{Informativeness-Alignment} seeks to train the model to answer with the most informative and complete factual answers to the questions.  


Our contributions can be summarized as follows: 
\begin{enumerate}
    \item  We formulate a factual informativeness evaluation task, which is integrated into both the training and evaluation pipeline. Empirical results show that our method leads to more informative answers, validating the effectiveness of this evaluation formulation. 
    \item We propose a novel two-stage training framework called \emph{InFACT}, enhancing the model to generate more informative answers.
    \item We evaluate our proposed framework with different models and demonstrate that it is effective in both improving information and factual accuracy. Our findings show significant improvement in factual precision, suggesting that the model learned to refrain from answering questions it would have made mistakes on, while overall high factual recall is preserved despite minor drops.
    \item Finally, we conduct in-depth analyses of both methods individually and potential spurious correlations. Our results confirm that the proposed framework outperforms several strong baselines focused on factual accuracy alone, enhancing both informativeness and factual reliability simultaneously. 
\end{enumerate}

\section{Informativeness Problem Setup}
\label{Sec:problem_setup}


\subsection{Background}

We first rigorously define a general setup of the \textit{informativeness evaluation task} which is the foundation of our alignment model. In factual question answering, each question $q_i$, e.g., \circled{1}\emph{What is the location of the capital of Australia?''} has a correct answer $t_i$ (e.g., \emph{the city of Canberra''}), which may appear in different string representations but refers to the same real-world entity. However, for a certain factual question $q_i$, there might exist several different correct answers. We distinguish the following three scenarios (see Fig.~\ref{figure:qa_types} for details):
\begin{enumerate}
    \item \textbf{Multiple-Answer Questions.} A question with multiple correct answers -- such as the question \emph{``What are the awards received by Barack Obama?''}, which has several correct answers referring to different entities.
    \item \textbf{Descriptive QA}. Responses may differ in their verbosity, yet point to the same real-world entity. For example -- the text \emph{``The city is located inland, about 150 kms from the coast, and is surrounded by picturesque mountains ...``} is the correct answer to \circled{1}. 
    
    \item \textbf{Granularity-based QA}. Answers to certain questions may have a varying level of completeness yet be factually correct; for instance, the valid answers to the question \emph{``Where was Barack Obama born?''}, would be \emph{``Kapiolani Medical Center for Women \& Children''}; or \emph{``Honolulu''} where the hospital is located; or \emph{``United States''} where the city is located; or all of them, depending on the level of granularity.

\end{enumerate}

\noindent Considering these types of QA, we formulate the dataset and the evaluation. 

\subsection{Dataset Formulation}
\label{sec:dataset_formulation}

An informativeness evaluation question-answering dataset is composed of a collection of questions, where we define a hierarchy of factually correct answers with varying levels of informativeness for each question. 
For a given question $q_i$, Level 1 includes the most \emph{informative} answers, while $L_i$ has the least informative ones; $L_i$ being the number of levels in the hierarchy of answers. 
More formally, we define a general informativeness-evaluation dataset as $\mathcal{D} = \{(q_i, H_i)\}_{i = 1}^{N}$, where $N$ is the number of questions in the dataset, $q_i$ is the $i$-th question in the dataset, and $H_i = (A_1, \dots, A_{L_i})$ is the corresponding hierarchy of answers.
A given $A_j = \{a_1, ..., a_{b_j}\}$ represents the $j$-th level in such a hierarchy, which is a set of all the possible answers at this level. For example, in Figure~\ref{figure:intro}, the answer \emph{Kapiolani Medical Center for Women \& Children, Honolulu, Hawaii, United States} is considered to be in set $A_1$, \emph{Honolulu, Hawaii, United States} in $A_2$, and \emph{United States} in $A_3$, reflecting the descending order of informativeness.

\section{Informativeness Alignment}

In this section, we present a detailed overview of our proposed method, \emph{InFACT}, which leverages the informativeness-evaluation dataset introduced in the last section to improve both the informativeness and factual correctness of LLM outputs.

\subsection{Baseline Mechanism}
\label{sec:baseline}

In the baseline model, we select the most informative set of answers, for instance $A_1$, as the gold label for each question, and fine-tune the LLM  using a randomly chosen answer from the selected set $A_1$. The goal is to train the model to learn the structure, preferring the most informative answer, such that it follows this structure during testing.




\subsection{Abstention Detection}
\label{sec:abstention_detection}

Prior to our Informativeness-Training (see Section~\ref{sec:informativness-training}), we introduce an automated LLM agnostic abstention detection mechanism to detect when an LLM abstains from answering a question. This is needed because when the LLM abstains from answering instead of answering incorrectly, we provide a higher reward for the model, thereby enhancing factual reliability. We define a function that accepts natural language text and determines whether the LLM output is any form of abstention from answering.  In this work, we leveraged \emph{GPT-4}, a powerful LLM in an in-context learning setup. A wide range of in-context data for abstention detection is collected from the LLM outputs of our experiments. 
For a given text $x$, we consider $x \in \textsc{Abstain}$, if and only if our abstention detection mechanism classifies it as a text that is in abstaining form. This mechanism encourages the model to prefer to err on the side of caution.

\subsection{Structure Tuning}
\label{sec:structure_tuning}

For efficient training, we first tune the model for the following:

\begin{enumerate}
    \item Optimizing the model to consistently generate the most factual and informative answers
    \item Learning to abstain more accurately, which most of the existing models lack without any additional training, often generating misinformation instead. 
\end{enumerate}

\noindent Therefore, prior to the informativeness-tuning phase, in structure tuning, we conduct the following training: For every question $q_i$ in our training set, we first let the model generate its answer. If the answer is correct, namely, it appears in one of the levels of $H_i$, then we train the model with a random answer from the level above it, i.e., the model is encouraged to learn a more informative answer. Otherwise, if the answer is in the highest level, i.e., it is the most informative answer according to our dataset, we take no further action, as the model has achieved the desired outcome. If the answer is incorrect, we teach the model to abstain. More formally -- for a model $M$ and a question $q_i$, the gold label, $L_\text{G}(M, q_i)$, is defined by: 

\begin{gather}
\label{eqn:gold_label_def}
{
   L_\text{G}(M, q_i) = \begin{cases} 
        \textsc{Random}(A_{j-1}) & \text{if } \exists j\geq 2: \hat{y} \in A_j\\
        \textsc{Skip}  & \text{if } \hat{y} \in A_1 \\
        \textsc{IDK} & \text{otherwise.} 
    \end{cases}%
\raisetag{2.2\baselineskip}} 
\end{gather}

\noindent Here, \textsc{IDK} denotes a response such as {``I don't know the answer''}.

\subsection{Informativeness Alignment}
\label{sec:informativness-training}

\normalsize
To fully leverage the hierarchical structure of our dataset, we introduce a preference-based training mechanism to reward the model for a more informative and correct answer. Additionally, we enhance abstention for potentially incorrect or misleading LLM outputs. As mentioned earlier, our dataset consists of a hierarchy of answers for each question, which is exploited to design a reward function to train an LLM as a policy to maximize its reward. 
Recall that for each question $q_i$ in our dataset, there is a corresponding hierarchy of answers $H_i$ as defined earlier.
Let $M$ be our LLM and $M(q_i) = \hat{y}$ be the model's answer to the question $q_i$. We define the following reward function: 
\vspace*{-3mm}
\begin{equation}
    \label{eqn:reward_func}
    R(M, \hat{y}) = \begin{cases} 
\frac{1}{\sqrt{j}} & \text{if } \exists j: \hat{y} \in A_j \\
 0 & \text{if } \hat{y} \in \textsc{Abstain}\\
 -1 & \text{otherwise.} 
\end{cases} 
\end{equation}

\noindent Observe that as long as there exists $j$ such that $\hat{y} \in A_j$, this means that $\hat{y}$ is correct, and thus we provide the model with a positive reward. The magnitude of the reward depends on the level in the hierarchy in which this correct answer is located; i.e., the more informative the answer, the larger the reward. In cases of the model output being wrong, meaning, its answer $\hat{y}$ is not at any level of the hierarchy, the model receives a negative reward, i.e., it is penalized to prevent hallucinations and factual errors. 

Having defined our reward function, we use RL techniques -- specifically PPO -- to train an LLM as a policy to maximize the reward. We also consider the reward function as a preference score and use a preference-training algorithm, and we use DPO in our experiments. This approach effectively balances completeness and correctness, resulting in more informative and factually reliable outputs.

\subsection{Overall Framework}


To summarize, our overall training mechanism of \emph{InFACT} is as follows: 
\circled{1} \textbf{Initialization}: We begin with any foundation model, which can be either a pre-trained or an instruction-tuned LLM. \circled{2} We deploy our \textbf{Structure-Tuning} mechanism (Section~\ref{sec:structure_tuning}) to extract relevant information and lay the ground work for the informativeness-alignment training. \circled{3} Finally, we introduce the second training phase, the \textbf{informativeness-alignment} technique (Section~\ref{sec:informativness-training}) to optimize the model to generate more informative and factually correct outputs.

\section{Experimental Setup}

In this section, we describe our experimental setup, including the investigated models, our baselines, and experimental details of our training procedure. 
    
\paragraph{Models.}

In our experiments, for both the baseline and the full completeness training, we use the following models: \llamaoneb{}, \llamathreeb{}, \llamaeightb{}, and \llamaeightbinstruct{} \cite{touvron2023llama, dubey2024llama}, \mistralsevenb{} \cite{Jiang2023Mistral7} and \qwensevenb{} \citep{Bai2023QwenTR, yang2024qwen2}.

\paragraph{Training Data.}
\label{sec:training_data}

\begin{table}
\centering
\renewcommand{\arraystretch}{0.8}
\begin{tabular}{@{}lp{1.2cm}@{}}
\toprule
\textbf{Statistic} & \textbf{Value} \\
\midrule
Total number of examples & 3,000 \\
Average number of levels per example & 8.9 \\
Average number of answers per level & 4.6 \\
\bottomrule
\end{tabular}
\caption{Summary statistics of the dataset.}
\label{table:dataset_statistics}
\end{table}
In order to create our training dataset, we follow the setup in Section~\ref{Sec:problem_setup}, as well as the formulation in Section~\ref{sec:dataset_formulation}. To construct our training set, we use the following datasets: GRANOLA QA \cite{yona-etal-2024-narrowing}, QAMPARI \cite{amouyal-etal-2023-qampari}, and RoMEQA \cite{Zhong2022RoMQAAB}. Specifically,  we randomly sample 1k examples from each. The examples from GRANOLA QA are already organized in the way we defined in Section~\ref{sec:dataset_formulation}, i.e., as a hierarchy of answers consisting of a sequence of answer levels. For the examples from QAMPARI and RoMEQA, we adjust the structure in the following way:
In each of these two datasets, each question has a list of answers. We construct our hierarchy with all single answers as the gold answers in the lowest level, followed by all possible pairs, triplets, and so on — culminating in the full list of answers with all permutations as the most informative response. 
Table~\ref{table:dataset_statistics} shows the statistics about the dataset. The main takeaway is the average number of levels per example, which is 8.9, as demonstrating that each of the questions is associated with a broad range of different informative answers.
Table~\ref{table:dataset_example_2} in Appendix~\ref{sec:training_details} presents two sample examples from the dataset, along with the corresponding training details provided in the same section.



\paragraph{Evaluation Data.}

For informativeness evaluation, we consider the test split of the benchmarks we used to build our training data from (GRANOLA QA, QAMPARI and RoMEQA). For factual accuracy we consider several other QA datasets: TriviaQA \cite{joshi2017triviaqa}, PopQA \cite{mallen2022not}, TruthfulQA \cite{Lin2021TruthfulQAMH}, Natural Questions \cite{Kwiatkowski2019NaturalQA}, and PIQA \cite{Bisk2019PIQARA}. These cover a wide range of questions, for example, general knowledge trivia questions (TriviaQA), subject-relation-object facts phrased as questions (PopQA), real-world user queries (Natural Questions), questions about human falsehoods (TruthfulQA), and physical commonsense reasoning questions (PIQA).
We consider the closed-book open-ended setting, where we do not provide any context or answer choices to the model.

\begin{table*}[t]
\setlength{\belowcaptionskip}{-4pt}
\centering 
\resizebox{1\linewidth}{!}{
\begin{tabular}{@{}l  cc  ccc  ccc@{}}
\toprule
&  \multicolumn{2}{c}{GRANOLA QA} & \multicolumn{3}{c}{QAMPARI} &  \multicolumn{3}{c}{RoMEQA} \\ 
\cmidrule(r){2-3}\cmidrule(lr){4-6}\cmidrule(lr){7-9}
\multicolumn{1}{c}{} & \textbf{Accuracy}  & \textbf{Informativeness}  & \textbf{P} & \textbf{R} & \textbf{F1}  & \textbf{P} & \textbf{R} & \textbf{F1}  \\
\midrule
\texttt{\llamaoneb}      & $45.1$    &$37.5$    
                    & $20.4$    &$3.1$   &$5.4$ 
                    & $24.4$    &$1.5$  &$2.8$     \\
\texttt{\llamaoneb}  + prompting    & $50.6$    &$39.1$    
                    & $17.7$    &$6.4$   &$9.4$ 
                    & $19.0$    &$2.9$  &$5.0$     \\
\texttt{\llamaoneb}  + FT    & $51.1$    &$45.2$    
                    & $20.6$    &$7.1$   &$10.6$ 
                    & $23.5$    &$4.9$  &$8.1$     \\
\textbf{\llamaoneb + informativeness-alignment}  & $\textbf{56.7}$    &$51.9$  
                    & $23.9$    &$\mathbf{10.8}$   &$\mathbf{14.9}$ 
                    & $\mathbf{25.9}$    &$10.4$   &$14.8$    \\ 
~~ + PPO      & $52.1$     &$\textbf{52.4}$   
                            & $ֿ\mathbf{24.7}$    &$10.7$   &$\mathbf{14.9}$
                            & $24.1$    &$\mathbf{11.2}$  &$\mathbf{15.3}$     \\  \midrule
\texttt{\llamathreeb}     & $51.3$     &$42.9$ 
                    & $\mathbf{23.1}$    &$4.3$ &$7.2$ 
                   & $23.6$    &$2.3$  &$4.2$ \\ 
\texttt{\llamathreeb}  + prompting    & $55.9$    &$46.0$    
                    & $19.6$    &$7.7$   &$11.0$ 
                    & $21.8$    &$6.7$  &$10.2$     \\
\texttt{\llamathreeb}  + FT    & $57.2$    &$50.3$    
                    & $23.0$    &$7.5$   &$11.3$ 
                    & $23.7$    &$7.1$  &$10.9$     \\
\textbf{\llamathreeb + informativeness-alignment}    & $\textbf{62.6}$     &$\textbf{56.8}$ 
                    & $22.4$    &$\mathbf{11.5}$ &$15.2$ 
                    & $\mathbf{23.8}$    &$\mathbf{11.8}$  &$\mathbf{15.8}$\\ \midrule
\texttt{\llamaeightb}    & $61.1$      &$53.4$ 
                    & $22.4$    &$8.9$ &$12.3$ 
                    & $30.4$    &$4.9$  &$15.8$  \\
\texttt{\llamaeightb}  + prompting    & $71.2$    &$56.8$    
                    & $21.5$    &$9.5$   &$13.2$ 
                    & $\mathbf{32.5}$    &$6.6$  &$10.2$     \\
\texttt{\llamaeightb}  + ICL    & $63.1$    &$58.5$    
                    & $22.7$    &$13.9$   &$17.2$ 
                    & $32.4$    &$8.6$  &$13.6$     \\
\texttt{\llamaeightb}  + FT    & $71.5$    &$58.0$    
                    & $22.5$    &$12.9$   &$16.4$ 
                    & $32.5$    &$13.1$  &$18.7$     \\
\textbf{\llamaeightb + informativeness-alignment} & $\mathbf{74.8}$       &$\mathbf{64.2}$
                            & $\mathbf{22.7}$    &$\mathbf{17.4}$   &$\mathbf{19.7}$   
                            & $\mathbf{32.5}$    &$\mathbf{13.1}$  &$\mathbf{18.7}$     \\
~~ - Structure-Tuning     & $64.5$     &$61.1$   
                            & $21.3$    &$11.8$   &$15.2$
                            & $26.7$    &$8.3$   &$12.7$     \\
~~ - Informativeness-Alignment      & $64.8$     &$57.4$   
                            & $21.5$    &$9.7$   &$13.4$
                            & $29.4$    &$6.4$  &$10.5$     \\   \midrule
\texttt{\mistralsevenb}     & $59.9$    &$53.9$    
                    & $21.0$    &$8.6$   &$12.2$ 
                    & $30.1$    &$5.5$  &$9.3$     \\
\texttt{\mistralsevenb}  + ICL    & $65.1$    &$60.0$    
                    & $22.8$    &$10.4$   &$14.3$ 
                    & $30.3$    &$7.9$  &$12.5$     \\
\texttt{\mistralsevenb}  + FT    & $62.5$    &$61.3$    
                    & $21.7$    &$11.9$   &$15.4$ 
                    & $29.6$    &$10.4$  &$15.4$     \\
\textbf{\mistralsevenb + informativeness-alignment}    & $\textbf{72.5}$     &$\textbf{64.1}$ 
                    & $22.9$    &$\mathbf{13.6}$ &$17.1$ 
                    & $\mathbf{30.8}$    &$\mathbf{12.9}$ 
                    &$\mathbf{18.2}$\\ \midrule
\texttt{\qwensevenb{}}     & $55.5$    &$51.7$    
                    & $22.4$    &$5.2$   &$8.4$ 
                    & $28.6$    &$4.9$  &$8.4$     \\
\texttt{\qwensevenb{}}  + ICL    & $56.5$    &$53.9$    
                    & $22.9$    &$5.8$   &$9.3$ 
                    & $28.5$    &$7.1$  &$11.4$     \\
\texttt{\qwensevenb{}}  + FT    & $56.4$    &$55.5$    
                    & $22.7$    &$6.9$   &$10.5$ 
                    & $28.4$    &$9.9$  &$14.7$     \\
\textbf{\qwensevenb{} + informativeness-alignment}    & $\textbf{68.8}$     &$\textbf{60.7}$ 
                    & $22.8$    &$\mathbf{11.5}$ &$15.3$ 
                    & $\mathbf{28.6}$    &$\mathbf{10.9}$ 
                    &$\mathbf{15.8}$\\ 
\bottomrule
\end{tabular}
}
\caption{Evaluation scores of \llamaoneb{}, \llamathreeb{}, \llamaeightb{}, \mistralsevenb{} and \qwensevenb{} on our informativeness evaluation QA benchmarks.}. 
\label{table:informativeness_results}
\end{table*}

\begin{table*}[t]
\setlength{\belowcaptionskip}{-4pt}
\centering
\resizebox{1\linewidth}{!}{
\begin{tabular}{@{}l  ccc  ccc  ccc  ccc  ccc@{}}
\toprule
&  \multicolumn{3}{c}{TriviaQA} & \multicolumn{3}{c}{PopQA} &  \multicolumn{3}{c}{TruthfulQA} & \multicolumn{3}{c}{Natural Questions} & \multicolumn{3}{c}{PIQA} \\ 
\cmidrule(r){2-4}\cmidrule(lr){5-7}\cmidrule(lr){8-10}\cmidrule(lr){11-13}\cmidrule(l){14-16}
\multicolumn{1}{c}{} & \textbf{P}  & \textbf{R} & \textbf{F1}  & \textbf{P} & \textbf{R} & \textbf{F1}  & \textbf{P}  & \textbf{R} & \textbf{F1} & \textbf{P}  & \textbf{R} & \textbf{F1} & \textbf{P}  & \textbf{R} & \textbf{F1} \\
\midrule
\texttt{\llamaoneb}      & $48.1$    &$48.1$   &$48.1$  
                    & $35.0$    &$35.0$   &$35.0$ 
                    & $29.5$    &$29.5$   &$29.5$ 
                    & $20.5$    &$20.5$   &$20.5$ 
                    & $71.9$    &$71.9$   &$71.9$  \\
\texttt{\llamaoneb} + Confidence Threshold    & $60.4$    &$38.3$   &$46.9$ 
                    & $42.9$    &$25.1$ &$31.6$ 
                    & $45.8$    &$14.9$ &$22.5$
                    & $33.0$    &$11.4$ &$16.9$
                    & $78.0$    &$55.8$ &$65.0$  \\ 
\texttt{\llamaoneb} + prompting  & $51.8$    &$40.8$   &$45.6$  
                    & $34.1$    &$22.8$   &$27.3$ 
                    & $39.4$    &$21.0$   &$27.4$ 
                    & $35.5$    &$14.9$   &$20.1$ 
                    & $76.9$    &$59.6$   &$67.1$  \\
\texttt{\llamaoneb} + P(True)    & $54.4$    &$40.5$   &$46.4$ 
                    & $39.6$    &$22.0$ &$28.3$ 
                    & $44.4$    &$20.7$ &$30.0$
                    & $36.2$    &$15.5$ &$21.7$
                    & $79.3$    &$60.1$ &$68.4$  \\ 
\texttt{\llamaoneb} + Semantic Entropy    & $57.8$    &$41.0$   &$48.0$ 
                    & $41.2$    &$21.9$ &$28.6$ 
                    & $44.9$    &$21.1$ &$28.7$
                    & $35.3$    &$18.4$ &$24.2$
                    & $78.5$    &$65.1$ &$71.2$  \\ \midrule
\texttt{\llamaoneb} + FT    & $49.1$    &$46.5$   &$47.8$ 
                    & $32.4$    &$37.0$ &$34.5$ 
                    & $40.1$    &$24.9$ &$30.7$
                    & $20.4$    &$23.8$ &$22.0$
                    & $73.5$    &$69.8$ &$71.6$  \\ \midrule
\textbf{\llamaoneb + informativeness-alignment} & $\mathbf{63.8}$    &$40.6$   &$\mathbf{49.6}$
                            & $\mathbf{47.6}$    &$31.0$   &$37.5$   
                            & $49.8$    &$24.5$   &$32.8$
                            & $\mathbf{35.9}$    &$20.0$   &$\mathbf{25.7}$
                            & $\mathbf{78.1}$    &$70.5$   &$\mathbf{74.1}$   \\
~~ + PPO      & $61.5$    &$43.0$   &$50.6$   
                            & $47.2$    &$31.5$   &$\mathbf{37.8}$
                            & $\mathbf{50.1}$    &$24.8$   &$\mathbf{33.2}$
                            & $34.1$    &$20.4$   &$25.5$
                            & $76.6$    &$71.2$   &$73.8$    \\
~~ - Structure-Tuning      & $54.2$    &$41.2$   &$46.8$   
                            & $40.5$    &$30.7$   &$34.9$
                            & $43.2$    &$23.8$   &$30.7$
                            & $32.9$    &$20.2$   &$25.0$
                            & $73.2$    &$70.8$   &$72.0$    \\
~~ - Informativeness-Alignment      & $54.2$    &$31.0$   &$39.4$   
                            & $49.8$    &$24.5$   &$32.8$
                            & $35.9$    &$20.0$   &$25.7$
                            & $31.9$    &$18.1$   &$23.1$
                            & $74.6$    &$67.8$   &$71.0$    \\

\bottomrule
\end{tabular}
}
\caption{Precision (P), Recall (R), and F1-scores for \llamaoneb{}. Our informativeness-aligned model achieves the best precision with minor decreases in recall, outperforming previous work.}. 
\label{table:llama_1b_factuality}
\end{table*}

\paragraph{Baselines.}

For informativeness evaluation, we compare the \tuned{} model with its original base model without any further training.
For factuality evaluation, we proceed similarly, but consider three different baselines:

\begin{enumerate}
    \item \textbf{Confidence Threshold} baseline: We use the predicted probability of the first generated token from the LM’s modeling head as a confidence score, following \citet{yoshikawa-okazaki-2023-selective}.
    If this confidence score is greater than a fixed threshold, we consider it as a valid generation; otherwise, we consider this as an uncertainty expression (analogous to abstention in our model). To create a strong baseline, we find the best threshold via hyperparameter tuning on the development set.

    \item \textbf{Prompting} baseline: We adopt a zero-shot approach where we instruct the model to be more informative in its answers but also to abstain in cases it does not know the answer. We use the following prompt: \emph{``Please answer the following question. Please answer with the most informative answer you can. Please refrain form answering if you don't know the correct answer. The question is: ''}. 

    \item \textbf{ICL} baseline: We adopt a few-shot approach by implicitly instructing the model to be more informative in its answers, using in-context demonstrations. Specifically, we sample 8 examples from our training set, and use the most informative answers as gold answers.

    \item \textbf{P(True)} baseline \citep{Kadavath2022LanguageM}: Given an input sentence to complete, say $I$, we use the original model to generate the completion, $A$. We then concatenate $I$ and $A$ and ask the model: \emph{``Please answer either with `true' or `false' only. Is it true that: $I A$''}. If the model answer is not `true', we consider this specific example as unknown for the model -- similar to our model abstaining.

    \item \textbf{Semantic Entropy} baseline \citep{kuhn2023semantic, aichberger2024semantically}: We sample $K$ text generations from the model encoding them using a state-of-the-art semantic encoder and cluster their encodings. If the largest cluster size is larger than $\frac{K}{2}$, we take a random generation from this cluster as the model's answer; otherwise, consider it as an unknown. 

    \item \textbf{FT Baseline}: This is described in Section~\ref{sec:baseline}. 
    This baseline evaluation distinguishes the effects of our proposed informativeness-alignment method and the created training set. 
\end{enumerate}

\paragraph{Evaluation.}
We assess the effect of our proposed model by measuring its \textit{informativeness, factuality, and knowledge retention}. For informativeness, we adopt the metrics that have been used in the original benchmarks. For example, GRANOLA QA provides a specific evaluation metric for informativeness that takes into account the location of the model's answer in the hierarchy of answers. QAMPARI and RoMEQA measure the precision and recall of the list of answers given by the model. For factuality and knowledge recall, we use the following metrics: \circled{1} \textbf{Precision}: the portion of factually correct answers out of all the questions that have a non-abstaining answer determined by our abstention detection model (see Section~\ref{sec:abstention_detection}), i.e., the questions that the model was certain about, and tried to be factually complete. \circled{2} \textbf{Recall}: the portion of factually correct answers out of all the questions in the dataset, namely, the portion of knowledge retention the model has, out of the entire test set.
\circled{3} \textbf{F1}: the harmonic mean of precision and recall. In the case of base models without additional calibration methods, the precision, recall, and F1-scores all correspond to their accuracy.

\begin{table*}[t]
\setlength{\belowcaptionskip}{-4pt}
\centering
\resizebox{1\linewidth}{!}{
\begin{tabular}{@{}l  ccc  ccc  ccc  ccc  ccc@{}}
\toprule
&  \multicolumn{3}{c}{TriviaQA} & \multicolumn{3}{c}{PopQA} &  \multicolumn{3}{c}{TruthfulQA} & \multicolumn{3}{c}{Natural Questions} & \multicolumn{3}{c}{PIQA} \\ 
\cmidrule(r){2-4}\cmidrule(lr){5-7}\cmidrule(lr){8-10}\cmidrule(lr){11-13}\cmidrule(l){14-16}
\multicolumn{1}{c}{} & \textbf{P}  & \textbf{R} & \textbf{F1}  & \textbf{P} & \textbf{R} & \textbf{F1}  & \textbf{P}  & \textbf{R} & \textbf{F1} & \textbf{P}  & \textbf{R} & \textbf{F1} & \textbf{P}  & \textbf{R} & \textbf{F1} \\
\midrule
\texttt{\llamathreeb}      & $50.2$    &$50.2$   &$50.2$  
                    & $36.1$    &$36.1$   &$36.1$ 
                    & $32.4$    &$32.4$   &$32.4$ 
                    & $26.0$    &$26.0$   &$26.0$ 
                    & $75.4$    &$75.4$   &$75.4$  \\
\texttt{\llamathreeb} + Confidence Threshold    & $58.9$    &$40.1$   &$47.7$ 
                    & $37.2$    &$29.2$ &$32.7$ 
                    & $34.8$    &$25.8$ &$29.6$
                    & $34.6$    &$20.0$ &$25.3$
                    & $78.5$    &$66.7$ &$72.1$  \\ 
\texttt{\llamathreeb} + prompting  & $52.3$    &$43.9$   &$47.7$  
                    & $36.2$    &$33.4$   &$34.7$ 
                    & $32.1$    &$32.6$   &$32.3$ 
                    & $29.7$    &$25.4$   &$27.4$ 
                    & $79.0$    &$68.2$   &$73.2$  \\
\texttt{\llamathreeb} + P(True)    & $58.2$    &$42.5$   &$49.1$ 
                    & $36.7$    &$31.0$ &$33.6$ 
                    & $35.5$    &$26.1$ &$30.1$
                    & $34.4$    &$24.9$ &$28.9$
                    & $76.6$    &$69.7$ &$73.0$  \\ 
\texttt{\llamathreeb} + Semantic Entropy    & $56.5$    &$44.1$   &$49.5$ 
                    & $39.0$    &$29.7$ &$33.7$ 
                    & $38.9$    &$25.7$ &$30.1$
                    & $34.0$    &$25.3$ &$29.0$
                    & $76.6$    &$71.4$ &$73.7$  \\ \midrule
\texttt{\llamathreeb} + FT    & $47.2$    &$47.2$   &$47.2$ 
                    & $39.5$    &$36.7$ &$38.0$ 
                    & $36.6$    &$30.8$ &$33.4$ 
                    & $27.8$    &$27.2$ &$27.5$ 
                    & $77.5$    &$74.9$ &$76.1$   \\ \midrule
\textbf{\llamathreeb + informativeness-alignment} & $\mathbf{64.0}$    &$42.9$   &$\mathbf{51.4}$
                            & $\mathbf{44.0}$    &$33.7$   &$\mathbf{38.2}$   
                            & $\mathbf{45.5}$    &$29.8$   &$\mathbf{36.0}$
                            & $\mathbf{48.1}$    &$25.4$   &$\mathbf{33.2}$
                            & $\mathbf{83.2}$    &$72.5$   &$\mathbf{77.5}$   \\
~~ - Structure-Tuning      & $55.7$    &$43.1$   &$48.6$   
                            & $36.9$    &$34.4$   &$35.6$
                            & $37.4$    &$28.4$   &$32.3$
                        
                            & $33.3$    &$25.0$   &$28.6$
                            & $75.2$    &$73.4$   &$74.3$    \\
~~ - Informativeness-Alignment      & $56.8$    &$40.3$   &$47.8$   
                            & $37.6$    &$30.1$   &$33.4$
                            & $37.2$    &$24.6$   &$29.6$
                            & $26.9$    &$24.3$   &$29.3$
                            & $76.8$    &$67.6$   &$71.4$    \\

\bottomrule
\end{tabular}
}
\caption{Precision (P), Recall (R), and F1-scores for \llamathreeb{}. Our informativeness-aligned model achieves the best precision with minor decreases in recall, outperforming previous work.}
\label{table:llama_3b_factuality}
\end{table*}

\begin{table*}[t]
\setlength{\belowcaptionskip}{-4pt}
\centering
\resizebox{1\linewidth}{!}{
\begin{tabular}{@{}l  ccc  ccc  ccc  ccc  ccc@{}}
\toprule
&  \multicolumn{3}{c}{TriviaQA} & \multicolumn{3}{c}{PopQA} &  \multicolumn{3}{c}{TruthfulQA} & \multicolumn{3}{c}{Natural Questions} & \multicolumn{3}{c}{PIQA} \\ 
\cmidrule(r){2-4}\cmidrule(lr){5-7}\cmidrule(lr){8-10}\cmidrule(lr){11-13}\cmidrule(l){14-16}
\multicolumn{1}{c}{} & \textbf{P}  & \textbf{R} & \textbf{F1}  & \textbf{P} & \textbf{R} & \textbf{F1}  & \textbf{P}  & \textbf{R} & \textbf{F1} & \textbf{P}  & \textbf{R} & \textbf{F1} & \textbf{P}  & \textbf{R} & \textbf{F1} \\
\midrule
\texttt{\llamaeightb}      & $53.5$    &$53.5$   &$53.5$  
                    & $38.5$    &$38.5$   &$38.5$ 
                    & $34.9$    &$34.9$   &$34.9$ 
                    & $28.9$    &$28.9$   &$28.9$ 
                    & $80.2$    &$80.2$   &$80.2$  \\   
\texttt{\llamaeightb} + Confidence Threshold    & $61.6$    &$39.5$   &$48.1$ 
                    & $50.8$    &$31.2$ &$38.6$ 
                    & $55.6$    &$21.9$ &$31.4$
                    & $49.6$    &$18.7$ &$27.2$
                    & $84.8$    &$61.3$ &$71.2$  \\ 
\texttt{\llamaeightb} + prompting  & $64.2$    &$46.1$   &$53.7$  
                    & $51.4$    &$31.9$   &$39.4$ 
                    & $52.5$    &$26.8$   &$35.5$ 
                    & $46.4$    &$24.1$   &$31.7$ 
                    & $82.5$    &$67.4$   &$74.2$  \\  
\texttt{\llamaeightb} + ICL  & $65.3$    &$46.1$   &$54.0$  
                    & $58.4$    &$33.8$   &$42.8$   
                    & $52.4$    &$28.9$   &$37.2$   
                    & $55.7$    &$21.4$   &$30.9$   
                    & $83.4$    &$67.8$   &$74.8$    \\ 
\texttt{\llamaeightb} + P(True)    & $65.4$    &$44.5$   &$53.0$ 
                    & $54.9$    &$30.7$ &$39.4$ 
                    & $55.2$    &$23.7$ &$33.2$
                    & $50.2$    &$21.5$ &$30.1$
                    & $82.2$    &$68.5$ &$74.7$  \\ 
\texttt{\llamaeightb} + Semantic Entropy    & $64.5$    &$45.8$   &$53.6$ 
                    & $55.5$    &$31.4$ &$40.1$ 
                    & $50.9$    &$24.6$ &$33.2$
                    & $51.2$    &$22.3$ &$31.1$
                    & $79.1$    &$68.9$ &$73.6$  \\ \midrule
\texttt{\llamaeightb} + FT    & $54.7$    &$54.7$   &$54.7$ 
                    & $37.1$    &$37.1$ &$37.1$ 
                    & $40.1$    &$32.1$ &$35.6$ 
                    & $27.2$    &$29.8$ &$28.4$ 
                    & $79.9$    &$80.2$ &$80.0$   \\ \midrule
\textbf{\llamaeightb + informativeness-alignment} & $\mathbf{70.1}$    &$47.2$   &$\mathbf{56.4}$
                & $\mathbf{65.5}$    &$35.9$   &$\mathbf{46.4}$   
                & $\mathbf{59.4}$    &$26.2$   &$\mathbf{36.4}$
                & $\mathbf{61.7}$    &$21.5$   &$\mathbf{31.8}$
                 & $\mathbf{86.9}$    &$75.1$   &$\mathbf{80.6}$   \\
~~ - Structure-Tuning      & $59.5$    &$49.1$   &$53.8$   
                            & $58.0$    &$35.6$   &$44.1$
                            & $51.3$    &$26.0$   &$32.7$
                            & $50.0$    &$21.8$   &$30.4$
                            & $81.0$    &$68.5$   &$74.2$    \\
~~ - Informativeness-Alignment      & $57.1$    &$49.9$   &$53.2$   
                            & $49.8$    &$35.1$   &$41.2$
                            & $46.8$    &$26.7$   &$34.0$
                            & $43.0$    &$22.1$   &$29.2$
                            & $84.1$    &$64.8$   &$73.2$    \\
\bottomrule
\end{tabular}
}
\caption{Precision (P), Recall (R), and F1-scores for \llamaeightb{}. Our informativeness-aligned model achieves the best precision with minor decreases in recall, outperforming previous work.}
\label{table:llama_8b_factuality}\end{table*}

\begin{table*}[t]
\setlength{\belowcaptionskip}{-4pt}
\centering
\resizebox{1\linewidth}{!}{
\begin{tabular}{@{}l  ccc  ccc  ccc  ccc  ccc@{}}
\toprule
&  \multicolumn{3}{c}{TriviaQA} & \multicolumn{3}{c}{PopQA} &  \multicolumn{3}{c}{TruthfulQA} & \multicolumn{3}{c}{Natural Questions} & \multicolumn{3}{c}{PIQA} \\ 
\cmidrule(r){2-4}\cmidrule(lr){5-7}\cmidrule(lr){8-10}\cmidrule(lr){11-13}\cmidrule(l){14-16}
\multicolumn{1}{c}{} & \textbf{P}  & \textbf{R} & \textbf{F1}  & \textbf{P} & \textbf{R} & \textbf{F1}  & \textbf{P}  & \textbf{R} & \textbf{F1} & \textbf{P}  & \textbf{R} & \textbf{F1} & \textbf{P}  & \textbf{R} & \textbf{F1} \\
\midrule
\texttt{\mistralsevenb{}}      & $53.1$    &$53.1$   &$53.1$  
                    & $35.5$    &$35.5$   &$35.5$ 
                    & $33.2$    &$33.2$   &$33.2$ 
                    & $27.8$    &$27.8$   &$27.8$
                    & $77.8$    &$77.8$   &$77.8$  \\   
\texttt{\mistralsevenb{}} + prompting  & $60.5$    &$45.6$   &$52.0$  
                    & $47.8$    &$34.3$   &$39.9$ 
                    & $44.9$    &$30.1$   &$36.0$ 
                    & $48.7$    &$22.2$   &$30.5$
                    & $79.4$    &$68.9$   &$73.8$  \\
\texttt{\mistralsevenb{}} + ICL  & $48.5$    &$56.1$   &$52.0$  
                    & $56.9$    &$27.6$   & 
                    $37.2$ 
                    & $32.5$    &$34.9$   &$33.7$ 
                    & $50.5$    &$21.5$   &$30.2$
                    & $72.6$    &$79.5$   &$75.9$ \\   \midrule
\texttt{\mistralsevenb{}} + FT    & $57.6$    &$50.2$   &$53.6$  
                    & $42.0$    &$34.8$   &$38.0$ 
                    & $33.5$    &$31.1$   &$32.3$ 
                    & $32.8$    &$22.8$   &$26.9$
                    & $77.9$    &$76.8$   &$77.3$  \\  \midrule
\textbf{\mistralsevenb{} + informativeness-alignment} 
                    & $\mathbf{66.7}$    &$49.5$   &$\mathbf{56.8}$  
                    & $\mathbf{60.2}$    &$34.6$   &$\mathbf{43.9}$ 
                    & $\mathbf{48.8}$    &$30.9$   &$\mathbf{37.8}$ 
                    & $\mathbf{55.8}$    &$21.8$   &$\mathbf{31.4}$
                    & $\mathbf{83.6}$    &$74.4$   &$\mathbf{78.7}$  \\   
\bottomrule
\end{tabular}
}
\caption{Precision (P), Recall (R), and F1-scores for \mistralsevenb{}. Our informativeness-aligned model achieves the best precision with minor decreases in recall, outperforming previous work.}
\label{table:mistral_7b_factuality}\end{table*}


\section{Results}

We present the main results obtained in our experiments, focusing on testing the influence of our method on the factual accuracy and informativeness of the answers generated by the models.   

\subsection{Informativeness.}

To measure the informativeness dynamics post-training, we evaluate the model using the test set of the three different datasets used to build our training data: GRANOLA QA, QAMPARI, and RoMEQA, using the original metrics proposed for each dataset. Table~\ref{table:informativeness_results} summarizes the results. Evaluating Granola QA, the informativeness measurement clearly goes up, as does the recall of the answers on QAMPARI and RoMEQA. We thus can conclude that overall, our models demonstrate a substantial improvement at each of these benchmarks, suggesting that the aligned model has indeed captured the concept of informativeness and has improved in it. Moreover, comparing to our FT baseline, we observe substantial improvements in results demonstrating the significance of our alignment method in obtaining these improved results. 

\subsection{Factual Accuracy.}

Tables~\ref{table:llama_1b_factuality}, \ref{table:llama_3b_factuality}, \ref{table:llama_8b_factuality}, \ref{table:mistral_7b_factuality}, and \ref{table:qwen_7b_factuality} present the performance of our models, as well as the relevant baselines, on our QA benchmarks. As can be seen, across all model sizes and all benchmarks, the overall F1 performance of our models is the highest. These gains are mostly a result of the increase in precision, that is, the model has learned to generate significantly fewer incorrect answers and to refrain where appropriate. In addition, recall performance tends to not show any major drop, which implies that with high probability, a substantial portion of the knowledge remains preserved in the model's parameters after our training. Overall, our results are promising, as they demonstrate that the informativeness of the answers generated by the model has also increased - that is, we can improve both accuracy and informativeness without a tradeoff between the two.
Notably, similar to the informativeness results, our alignment method leads to a significant improvement in factual precision compared to the fine-tuned (FT) baseline, highlighting the impact of our approach.

\subsection{PPO vs.\ DPO}
As discussed in Section~\ref{sec:informativness-training}, to train our models, we can use both policy-optimization methods like PPO and preference-based algorithms such as DPO. Here we compare these two in terms of results. Tables~\ref{table:informativeness_results} and \ref{table:llama_1b_factuality} include the results of our model using the PPO algorithm, with \llamaoneb{}. As observed, the resulting gap between these two algorithms is not compelling enough to be statistically significant. We thus use only DPO for the larger models, due to resource considerations. 

\subsection{Ablation Study}

As outlined earlier, our method is composed of two components: Structure-Tuning (Section~\ref{sec:structure_tuning}) and Informativeness-alignment (Section~\ref{sec:informativness-training}). Here we study the specific impact of each of them separately. We follow the same experimental setup, yet we apply it to two different models: one only trained via structure-tuning and one trained via informativeness-alignment. Tables~\ref{table:informativeness_results}, \ref{table:llama_1b_factuality}, \ref{table:llama_3b_factuality}, and \ref{table:llama_8b_factuality} show these results for the different model sizes on the different evaluation datasets we have used. The pattern is evident -- omitting the structure-tuning degrades the factual precision of the resulting model. We trace that this decline stems from a lack of ability of the foundation models to effectively abstain, often generating misinformation.

\subsection{Error Analysis}
\label{sec:error_analysis}

To better assess our model's generations, we conducted two experiments: We randomly sampled 200 factual mistakes made by the model, and 40 from each of the factual evaluation datasets. We then compared the responses of our model with the original model, and our findings are as follows:

\begin{enumerate}[itemsep=0mm]
    \item For $84\%$ of the questions, the original model generated a wrong answer, which suggests that most of the mistakes are very likely due to the lack of parametric knowledge. 
    \item For $3\%$, the model abstained from answering, but our automatic mechanism did not recognize it. 
    \item For the remaining $13\%$, our models' answers were significantly longer (in terms of words). We assume that the model might have learned a spurious correlation from the proposed training that longer answers are more informative. 
\end{enumerate}

\section{Related Work}

\paragraph{Informativeness}

Evaluating the informativeness of dialogue agents and chatbots is a crucial research goal because of their widespread use. Some studies focus on evaluating the informativeness of dialogue model responses \cite{DeFreitas2020TowardsAH, thoppilan2022lamda, lu-etal-2023-towards}, and some on factual sentence completions \cite{Huang2022CanLM}. More recent work proposes a benchmark for evaluating the factual granularity of model responses \cite{yona-etal-2024-narrowing}, which we use in this work for both training and evaluation. External tools are also used for better informativeness and correctness \cite{schick2023toolformer}. In this work, we take a model alignment approach to tackle these open challenges.

\paragraph{Factuality} 

Factuality has been widely studied lately from various different perspectives \cite{augenstein2023factuality}. 
One way to tackle this is through the problem of factual error detection, where a binary prediction model is provided instead of a continuous probability.. This is also related to the setting of selective prediction, where models can abstain from answering a query \citep{varshney-etal-2022-investigating, kamath2020selective}. 
Another way to tackle this is via model calibration \citep{pmlr-v70-guo17a}. The goal is to provide a measure of the probability that a prediction is incorrect alongside the actual prediction. Common approaches to calibration are to perform various transformations on a model's output logits \citep{desai2020calibration, jiang-etal-2021-know}, and measuring uncertainty \citep[e.g., see][]{kuhn2023semantic}. More recent works have studied the use of LMs for providing calibration by training them on statements known to be factually correct or not. This ``supervised'' approach has been explored via fine-tuning \citep{Kadavath2022LanguageM, lin2022teaching}, in-context learning \citep{cohen-etal-2023-crawling, alivanistos2022prompting}, zero-shot instruction-oriented \citep{cohen-etal-2023-lm, dhuliawala2023chain, Feng2024DontHA}, and consistency sampling \citep{yoran-etal-2023-answering} techniques.
Further recent studies \citep{azaria-mitchell-2023-internal} use the internal state of the model for classifying whether it is certain or not, use a new token for unanswerable inputs \citep{lu-etal-2022-controlling} or for uncertainty representation \cite{cohen2025don}, or construct a specific dataset for effectively tuning the model for answering refusal \citep{zhang-etal-2024-r}. 
Our work addresses the factuality problem by aligning an LLM for better informativeness and correctness altogether.

\section{Conclusion}

We propose a novel method \emph{InFACT}, that aims to improve both the correctness and informativeness of LLM's responses. This mechanism takes advantage of factual questions which can be correctly answered at various levels of informativeness and aligns the LLM with more informative yet still correct answers.

An in-depth evaluation across diverse QA benchmarks suggests that this mechanism upgrades both the factual precision of the model's answers, by more effectively abstaining rather than generating wrong facts, as well as the informativeness of its answers, by generating a larger amount of correct answers (higher recall of multiple-answer questions), including  answers that are more descriptive and of more appropriate granularity.

This work has the potential to facilitate several intriguing follow-up studies. One of them is the curation of a new unified and qualitative dataset for informativeness evaluation, which may have the potential to further improve the factual consistency of the LLMs.

\section*{Limitations}

We note a few limitations of our method. First, it requires a decent amount of labeled data. This labeled data should also have a notion of informativeness, as our method requires an informativeness-based hierarchy of labels for each input example, as discussed in Section~\ref{sec:dataset_formulation}. For certain domains, this could be hard to obtain. In general, however, such data can be procured from numerous sources.  When existing datasets lack explicit answer hierarchies, they can be extended using external knowledge sources like KGs or crowd annotations. This highlights the flexibility of our framework, enabling the transformation of flat QA datasets into more informative training resources.

Second, as discussed in Section~\ref{sec:error_analysis}, the model might learn undesirable spurious correlations through the proposed alignment process, such as with the answer length, as the goal of the method is to teach the model to extract the most informative answer.  

Third, the design of our method was motivated by the assumption that one would rather obtain fully correct answers only. In this setting, it may occur that the model generates answers that are partially correct, but we then teach it to abstain instead. Thus, the method design as well as the evaluation might need to be customized to tailor to the needs in specific application setups or domains.

`

\bibliography{anthology,custom}
\bibliographystyle{acl_natbib}

\appendix

\section{Additional Results}
Here we attach essential additional experimental results.

\begin{table*}[t]
\setlength{\belowcaptionskip}{-4pt}
\centering
\resizebox{1\linewidth}{!}{
\begin{tabular}{@{}l  ccc  ccc  ccc  ccc  ccc@{}}
\toprule
&  \multicolumn{3}{c}{TriviaQA} & \multicolumn{3}{c}{PopQA} &  \multicolumn{3}{c}{TruthfulQA} & \multicolumn{3}{c}{Natural Questions} & \multicolumn{3}{c}{PIQA} \\ 
\cmidrule(r){2-4}\cmidrule(lr){5-7}\cmidrule(lr){8-10}\cmidrule(lr){11-13}\cmidrule(l){14-16}
\multicolumn{1}{c}{} & \textbf{P}  & \textbf{R} & \textbf{F1}  & \textbf{P} & \textbf{R} & \textbf{F1}  & \textbf{P}  & \textbf{R} & \textbf{F1} & \textbf{P}  & \textbf{R} & \textbf{F1} & \textbf{P}  & \textbf{R} & \textbf{F1} \\
\midrule
\texttt{\qwensevenb{}}      & $52.7$    &$52.7$   &$52.7$  
                    & $30.8$    &$30.8$   &$30.8$ 
                    & $30.2$    &$30.2$   &$30.2$ 
                    & $25.3$    &$25.3$   &$25.3$
                    & $78.6$    &$78.6$   &$78.6$  \\      
\texttt{\qwensevenb{}} + prompting  & $54.8$    &$50.5$   &$52.5$  
                    & $45.5$    &$30.4$   &$36.4$ 
                    & $42.1$    &$27.9$   &$33.5$ 
                    & $44.5$    &$19.7$   &$27.3$
                    & $80.4$    &$66.7$   &$72.9$  \\  
\texttt{\qwensevenb{}} + ICL  & $48.6$    &$53.7$   &$51.0$  
                    & $49.2$    &$25.8$   &$33.8$ 
                    & $27.0$    &$30.1$   &$28.5$ 
                    & $45.9$    &$22.2$   &$29.9$
                    & $72.4$    &$79.9$   &$76.0$  \\   \midrule
\texttt{\qwensevenb{}} + FT    & $52.0$    &$55.9$   &$53.8$  
                    & $31.8$    &$34.2$   &$33.0$ 
                    & $35.1$    &$34.5$   &$34.8$ 
                    & $30.3$    &$29.9$   &$30.1$
                    & $75.8$    &$79.9$   &$\mathbf{77.8}$  \\  \midrule
\textbf{\qwensevenb{} + informativeness-alignment} 
                    & $\mathbf{60.9}$    &$51.2$   &$\mathbf{55.6}$  
                    & $\mathbf{54.8}$    &$30.1$   &$\mathbf{38.9}$ 
                    & $\mathbf{49.1}$    &$27.6$   &$\mathbf{35.3}$ 
                    & $\mathbf{48.7}$    &$24.0$   &$\mathbf{32.2}$
                    & $\mathbf{80.5}$    &$72.7$   &$76.4$  \\   
\bottomrule
\end{tabular}
}
\caption{Precision (P), Recall (R), and F1-scores for \qwensevenb{}. Our informativeness-aligned model achieves the best precision with minor decreases in recall, outperforming previous work.}
\label{table:qwen_7b_factuality}\end{table*}

\section{Dataset Example}
Here we attache two examples from our training dataset \autoref{sec:training_data}. 

\begin{table}[h!]
\centering
\renewcommand{\arraystretch}{1.3}

\caption*{\textbf{Example (a)}: \textit{Where was Luke Prokopec born?}}
\begin{tabular}{@{}clp{7cm}@{}}
\toprule
\textbf{Level} & \textbf{Answer} & \textbf{Description} \\
\midrule
A1 & Blackwood & Specific town (most fine-grained answer) \\
A2 & Caerphilly County Borough & Local administrative district \\
A3 & Wales & Country within the UK \\
A4 & United Kingdom & Sovereign state (most general) \\
\bottomrule
\label{table:dataset_example_1}
\end{tabular}

\vspace{1.2cm}

\caption*{\textbf{Example (b)}: \textit{Where are newspapers owned at some point in time by Voice Media Group published?}}
\begin{tabular}{@{}clp{6cm}@{}}
\toprule
\textbf{Level} & \textbf{Answer(s)} & \textbf{Description} \\
\midrule
A4 & \{Houston, Dallas, Palm Beach, Phoenix, Denver\} & Individual cities (fine-grained locations) \\
A3 & \{"Houston and Dallas", "Phoenix and Denver", ...\} & Pairs of cities (small groupings) \\
A2 & \{"Dallas, Phoenix and Denver", ...\} & Medium-sized groupings of cities \\
A1 & \{"Houston, Dallas, Palm Beach, Phoenix and Denver"\} & All cities grouped as a single set \\
\bottomrule
\end{tabular}

\caption{Hierarchical answer representations for two different questions, from specific (A1) to more abstract (A4/A5).}
\label{table:dataset_example_2}
\end{table}

\section{Training Details}
\label{sec:training_details}

We use a maximum learning rate of $2 \times 10^{-5}$ with a linear warmup for 10\% of the training steps and a cosine decay down to $2 \times 10^{-6}$. We use a batch size of 256, weight decay of 0.05, gradient clipping of 1.0, and AdamW $\beta$ values (0.9, 0.95). We train for 256 optimization steps. Regarding infrastructure, we use 4 NVIDIA A100 40G GPUs. We use the same parameters and constraints for both stages - structure-tuning and informativeness-alignment.

\section{Different Types of QA data}
\label{sec:dataset_types}

 \begin{figure*}[ht]
\setlength{\belowcaptionskip}{-10pt}
    \centering
    \includegraphics[width=0.9\textwidth]{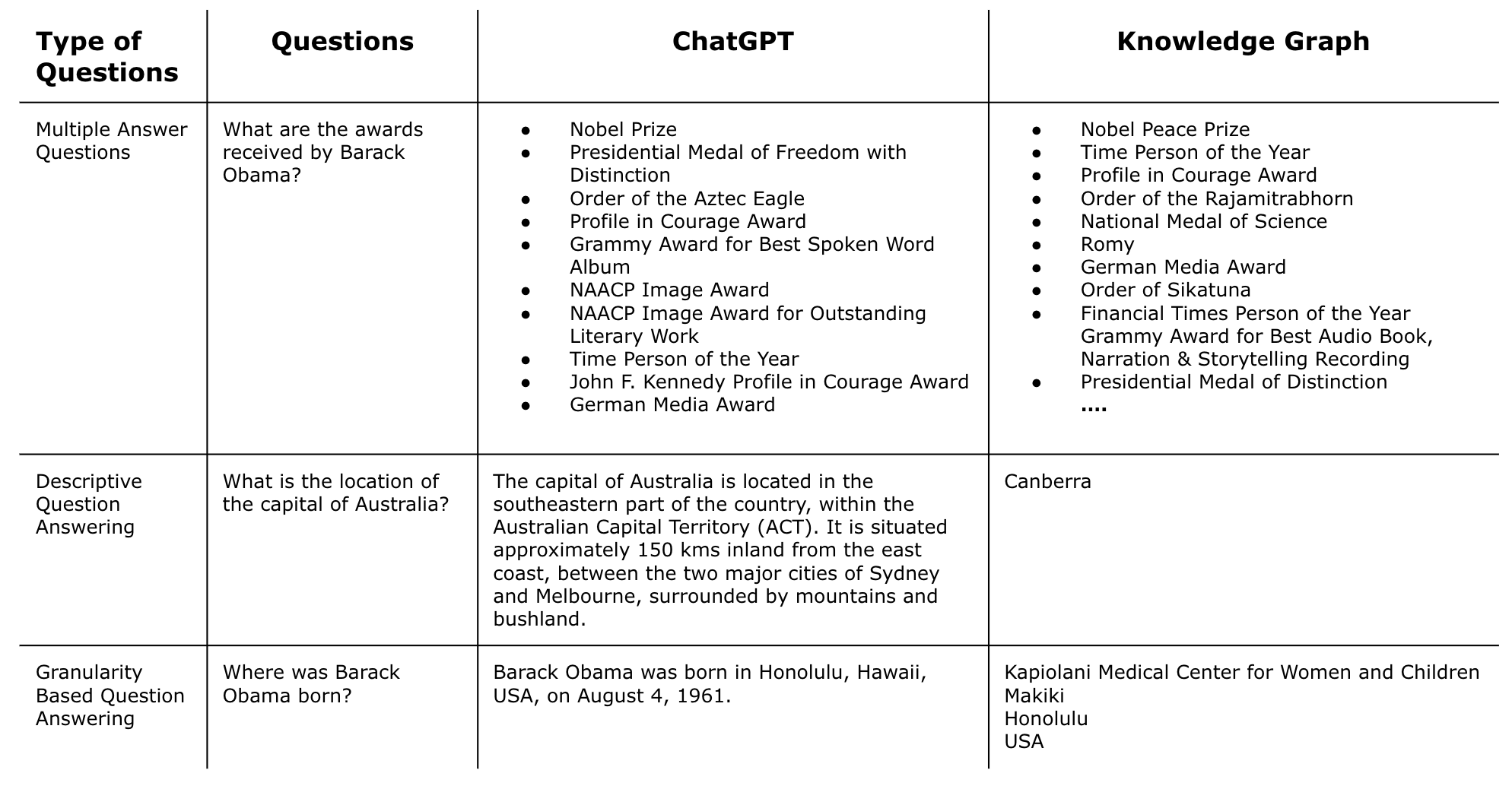}
     \caption{Different Types of Question Answers}
     \label{figure:qa_types}
\end{figure*}


\end{document}